\documentclass[10pt,twoside,twocolumn,a4paper]{article}

\usepackage[accepted]{bpasts}

\usepackage{t1enc}
\usepackage[utf8]{inputenc}
\usepackage[english]{babel}
\usepackage{amsmath,amsfonts,amssymb}

\usepackage{graphicx}
\usepackage[keeplastbox]{flushend}

\usepackage{url}            
\usepackage{relsize}        
\usepackage{booktabs,arydshln}       
\usepackage{nicefrac}       
\usepackage{microtype}      
\usepackage{pifont}         
\usepackage[hidelinks]{hyperref}       

\abtitle{Predicting Pairwise Relations with Neural Similarity Encoders}
\title{Predicting Pairwise Relations with Neural Similarity Encoders}

\abauthor{Franziska Horn \& Klaus-Robert Müller}
\author{Franziska Horn$^{1}$ and Klaus-Robert Müller$^{1,2,3}$\email{cod3licious@gmail.com}}

\Address{$^1$Machine Learning Group, Technische Universität Berlin, Berlin, Germany\\
$^2$Department of Brain and Cognitive Engineering, Korea University, Seoul, Republic of Korea\\
$^3$Max-Planck-Institut für Informatik, Saarbrücken, Germany}

\Abstract{Matrix factorization is at the heart of many machine learning algorithms, for example, dimensionality reduction (e.g.\ kernel PCA) or recommender systems relying on collaborative filtering. Understanding a singular value decomposition (SVD) of a matrix as a neural network optimization problem enables us to decompose large matrices efficiently while dealing naturally with missing values in the given matrix. But most importantly, it allows us to learn the connection between data points' feature vectors and the matrix containing information about their pairwise relations. In this paper we introduce a novel neural network architecture termed \emph{similarity encoder} (SimEc), which is designed to simultaneously factorize a given target matrix while also learning the mapping to project the data points' feature vectors into a similarity preserving embedding space. This makes it possible to, for example, easily compute out-of-sample solutions for new data points. Additionally, we demonstrate that SimEc can preserve non-metric similarities and even predict multiple pairwise relations between data points at once.}


\vol{66} \no{6} \year{2018}
\doi{10.24425/bpas.2018.125929}

\begin{document}
\maketitle


\section{Introduction}
Pairwise relations, such as similarities, between data points play an important role in many areas of machine learning (ML) \cite{bishop,hastie01statisticallearning,hofmann1997pairwise,scholkopf2002learning}. Dimensionality reduction methods such as t-SNE \cite{van2008visualizing}, kernel PCA (kPCA) \cite{scholkopf1998nonlinear}, isomap \cite{tenenbaum2000global}, and locally linear embedding (LLE) \cite{roweis2000nonlinear} create low dimensional representations of data points by preserving their pairwise similarities, distances, or local neighborhoods in the embedding space, e.g., to create informative visualizations of a dataset \cite{horn2017pubvis}. Similarity preserving embeddings of data points can also serve as useful feature representations for other (supervised) ML tasks. For example, by computing the eigendecomposition of a kernel (i.e.~similarity) matrix, kPCA projects the data into a feature space where data points can become linearly separable and noise in the data can be reduced \cite{mika1999kernel,muller2001introduction,scholkopf1999input}. In natural language processing (NLP) settings, the popular word2vec model \cite{mikolov2013efficient,mikolov2013distributed} learns an embedding for each word in the vocabulary by relying on the principle that similar words appear in similar contexts \cite{harris1954distributional,horn2017conecRepL4NLP,levy2014neural}. Using word embeddings as features can improve the performance in many NLP tasks such as named entity recognition or text classification \cite{collobert2011natural,le2014distributed,turian2010word}. The prediction of pairwise relations themselves is at the heart of important real world ML applications such as the prediction of whether or not a drug could interact with a certain protein \cite{gonen2012predicting} or for recommender systems, where the task is to predict the rating a user would give to a certain item \cite{koren2009matrix,sarwar2000application} or to identify similar items that could be promoted alongside an item of interest \cite{barkan2016item2vec}. Another active research area is concerned with the analysis of graphs, such as social networks, where the pairwise relations between nodes are of key importance \cite{ahmed2013distributed,hamilton2017representation}.

Such pairwise relations between data points can be represented as a rectangular matrix $R \in \mathbb{R}^{m\times n}$, which could, for example, contain the ratings of $m$ items by $n$ users. In the following, we will primarily focus on pairwise similarities between $m$ data points, stored in a square symmetric matrix $S \in \mathbb{R}^{m\times m}$, but also discuss how our results generalize to arbitrary pairwise relations $R$.

In this paper, we introduce our novel neural network architecture called \emph{similarity encoder} (SimEc), which learns (low dimensional) \emph{similarity preserving embeddings} for data points. To be more precise, SimEc represent the data in such a way that the scalar product between the embedding vectors $\mathbf{y}_i, \mathbf{y}_j \in \mathbb{R}^d$ of two data points approximates their similarity, i.e., $\mathbf{y}_i\mathbf{y}_j^\top \approx S_{ij}$. Furthermore, given some original (high dimensional) feature vectors $\mathbf{x}_i \in \mathbb{R}^D \; \forall i \in \{1,...,m\}$, a SimEc additionally provides a linear or non-linear mapping function $f': \mathbf{x}_i \to \mathbf{y}_i$, which can be used to project new data points into the similarity preserving embedding space, i.e., to compute out-of-sample (OOS) solutions. Here it is important to note that these feature vectors do not have to be directly related to the given target similarities stored in $S$, in fact, we do not need to know how $S$ was computed at all, e.g., it could also contain human similarity ratings. Furthermore, SimEc can deal with missing values in the similarity matrix $S$, can embed data points based on metric or non-metric similarities, and can be used to predict multiple pairwise similarities or other relations between data points at once. We provide a keras \cite{chollet2015keras} based Python implementation of the model, which can be trained efficiently on GPUs.\footnote{\url{https://github.com/cod3licious/simec}}

After relating our model to previous work, we detail its architecture in Section~\ref{sec:model} and then demonstrate its effectiveness in multiple experiments (Section~\ref{sec:exp}) before concluding the paper with a discussion.

\subsection{Related Work}
The optimal (in a least squares sense) low dimensional embeddings to factorize a matrix $R$ or $S$ can be found by computing a singular value decomposition (SVD) or eigendecompositon of the matrix and using the $d$ largest eigenvalues and corresponding eigenvectors to compute a low rank approximation of the matrix. However, performing an SVD is computationally very expensive for large matrices, and in these cases requires the use of approximate iterative methods \cite{koren2009matrix}. Furthermore, an exact decomposition can not be computed for matrices that contain missing values, in which case weighted error functions need to be employed \cite{hu2008collaborative}. Back in 1982, a simple neural network (NN) was conceived to compute a PCA \cite{oja1982simplified} and in 1992, NNs were proposed as a method to efficiently compute the SVD \cite{cichocki1992neuralsvd} or eigendecomposition \cite{cichocki1992neuralevd} of a matrix while naturally dealing with missing values in the target matrix, which we will discuss in more detail in Section~\ref{sec:nn_matrixfac}.

\begin{figure}[!ht]
  \centering
      \includegraphics[width=\linewidth]{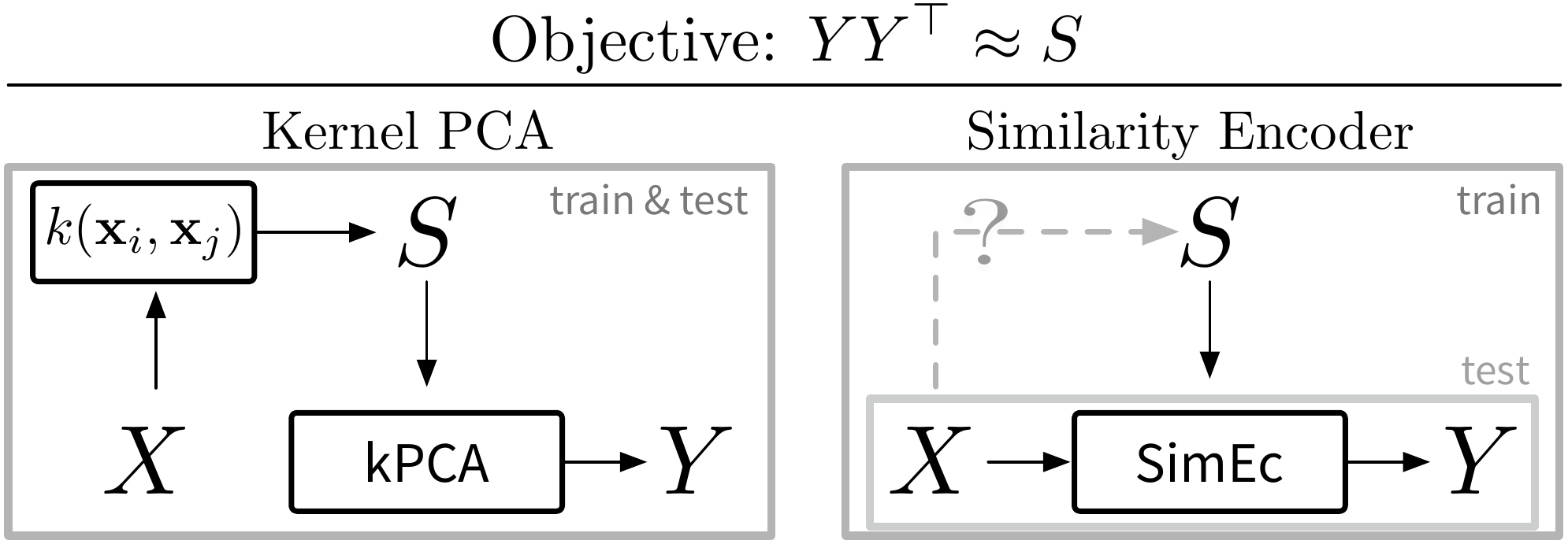}
  \caption{Kernel PCA and SimEc both aim to project the data points into an embedding space where the target similarities can be approximated by the scalar product of the embedding vectors, but kernel PCA also needs to compute a kernel map, i.e., the similarities to the training data points, to be able to embed new test samples.}
  \label{fig:simec_kpca_overview}
\end{figure}

If a similarity matrix was computed with a known support vector kernel function \cite{muller2001introduction,scholkopf2002learning,vapnik1995nature}, a manually devised, kernel-specific random mapping from the original input to the kernel feature space could be used to create similarity preserving embeddings for large datasets very efficiently \cite{rahimi2007random}. By interpreting this mapping as a neural network, it can be further fine-tuned to the dataset at hand \cite{alber2017empirical}.
But while spectral methods such as kPCA provide optimal similarity preserving embeddings based on the pairwise similarities for a given set of data points, a critical issue remains, namely that they can only compute OOS solutions, i.e., embeddings for new tests points, if their similarity to the original training examples can be computed with a known kernel function \cite{bengio2004out}. SimEc on the other hand learn a direct mapping from the original feature space to the similarity preserving embedding space and therefore do not require knowledge of how the pairwise similarities were computed (Fig.~\ref{fig:simec_kpca_overview}).
It would be possible to train an additional regression model to learn the mapping from the original input feature space to the embeddings computed by the spectral method \cite{mao1995artificial,teh2003automatic}. However, the best similarity preserving embeddings that can be realized as a transformation of the original feature vectors might not necessarily correspond to the embeddings found by the spectral method, thereby losing some information by the mapping, resulting in unnecessarily poor similarity approximations \cite{carreira2015fast}.\footnote{For example, eigenvalue $d$ might only be slightly larger than eigenvalue $d+1$, however, the $d$th eigenvector might contain information that is not present in the original feature vectors, while the information encoded in eigenvector $d+1$ can be preserved by a transformation of the feature vectors. By learning the mapping and factorization together, it is possible to create a $d$-dimensional embedding that instead retains the information from the $d+1$ eigenvalue and -vector, thereby resulting in an only slightly worse approximation of $S$ compared to the spectral method, while not losing any accuracy in the mapping step.}

Previous work concerned with simultaneously factorizing a matrix with pairwise relations while learning a mapping from the original input space to the low dimensional embedding space can be categorized into the two approaches outlined below. In both cases, the mapping from the input to the embedding space is usually realized by neural networks, as they provide the flexibility to learn arbitrary functions.
\subsubsection{Embedding with a single NN} In the first approach, a single neural network is trained to map the points into a similarity preserving embedding space by computing the embeddings for a batch of training samples and then comparing the pairwise similarities (or distances) of these embedded points against the target similarities to compute the error used in the backpropagation procedure to tune the network's parameters. With this approach, extensions for t-SNE \cite{maaten2009learning} and other classic manifold learning methods \cite{bunte2012general,lowe1996feed} were developed, which enable the computation of OOS solutions. A particularly interesting realization of this approach are deep kernelized autoencoders \cite{kampffmeyer2017deep}, which train an autoencoder network with an additional objective to not only minimize the reconstruction error of the data points themselves but also the mismatch between the dot product of a batch of embedding vectors and the corresponding block from a kernel matrix. The decoder part of the autoencoder network thereby also provides a mapping from the embedding space back to the original feature space, which can be used to compute the pre-image of an embedding vector \cite{mika1999kernel}.
By directly minimizing $\|S - YY^\top\|$, these methods successfully learn similarity preserving embeddings, however, because they always operate on batches of points, these methods scale quadratically and efficient training is highly dependent on the choice of the batch size, requiring either lots of memory or many combinations of randomly chosen samples to cover all pairwise similarities. In an effort to improve on this, the method of auxiliary coordinates can be used to train a NN in an alternating fashion, in one step optimizing the mapping from the input to the embedding space, in the other step improving the similarity approximation of the embedding itself \cite{carreira2015fast}. As we will see in Section~\ref{sec:simec}, while the weight matrix of the last layer of the SimEc architecture could be interpreted as a set of auxiliary coordinates as well, training a SimEc network does not require alternating steps in the optimization procedure.

The above mentioned methods all learn embeddings based on pairwise similarities, but can \emph{not} generalize to other types of pairwise relations, specifically those involving two different kinds of data, e.g., ratings of items by users.
\subsubsection{Learning two mappings into the same embedding space} This brings us to the second approach, where two networks are trained to simultaneously map two (different kinds of) input feature vectors into the same embedding space. Again, the objective is for the similarity between the two embedding vectors to approximate the respective pairwise relation stored in the target matrix $R$. For pairwise similarities, the mapping into the embedding space can also be realized by a siamese network, i.e., two networks with shared parameters operating on the same kinds of input data \cite{hadsell2006dimensionality}. As these networks operate on pairs of samples, training again scales quadratically, but here at least the batch sizes for both networks can be chosen independently. Furthermore, this approach is often employed for recommender systems \cite{huang2013learning}, where the target matrix $R$ is typically very very sparse, which means by training only on pairs of samples with known targets, the training time can be greatly reduced. In these cases, often some form of negative sampling is employed during training to consider for every positive sample pair (e.g.~a song a user has listened to) some negative pairs (songs a user has not listened to) as well, as these can provide additional information \cite{wu2017starspace}.
With a fast training procedure for target matrices where the number of non-zero elements is much smaller than $m\cdot n$ and the benefit of learning multiple mapping functions simultaneously for projecting different kinds of feature vectors into the same embedding space, this second approach is very useful for many applications scenarios. SimEc, on the other hand, are designed to efficiently factorize dense matrices (while being able to handle missing values in the target matrix) and, while they rely on only a single neural network to map input features into a similarity preserving embedding space, we discuss in the next section how they can be trained to \emph{predict multiple pairwise relations at once} based on the same embedding.

\section{The SimEc Model}\label{sec:model}
In the following, we will first describe how neural networks can realize the computation of an SVD of a rectangular matrix $R \in \mathbb{R}^{m\times n}$ \cite{cichocki1992neuralsvd} and the eigendecomposition of a square symmetric matrix $S \in \mathbb{R}^{m\times m}$ \cite{cichocki1992neuralevd}. Then we detail how these models can be extended to arrive at the SimEc neural network architecture.
\subsection{Matrix factorization with neural networks}\label{sec:nn_matrixfac}
With singular value decomposition (SVD), a matrix $R \in \mathbb{R}^{m\times n}$ can be decomposed as\vspace{-10pt}
\begin{align*}
R &= U \Sigma V^\top,
\end{align*}
where $U \in \mathbb{R}^{m\times m}$ and $V \in \mathbb{R}^{n\times n}$ contain the eigenvectors of $RR^\top$ and $R^\top R$ respectively while the corresponding eigenvalues are stored in $\Sigma \in \mathbb{R}^{m\times n}$. By using only the $d$ largest eigenvalues and corresponding eigenvectors, a low rank approximation of $R$ can be obtained, i.e.,\ $R \approx U_{[:,:d]} \Sigma_{[:d,:d]} V^\top_{[:d,:]}$.

By setting $W_1 = U_{[:,:d]} \sqrt{\Sigma_{[:d,:d]}}$ and $W_2 = \sqrt{\Sigma_{[:d,:d]}}V^\top_{[:d,:]}$, the low rank approximation of $R$ can be rewritten as
\begin{align*}
R &\approx W_1W_2 = I_mW_1W_2,
\end{align*}
where $I_m \in \mathbb{R}^{m\times m}$ is the identity matrix.

A simple feed forward neural network $f(\mathbf{x}_i)$ can now be constructed with two layers defined by the weight matrices $W_1 \in \mathbb{R}^{m\times d}$ and $W_2 \in \mathbb{R}^{d\times n}$ and without any non-linear activation functions. Given some input vector $\mathbf{x}_i \in \mathbb{R}^m$, the first layer computes\vspace{-10pt}
\begin{align*}
f'(\mathbf{x}_i) &= \mathbf{x}_i W_1 = \mathbf{y}_i,
\end{align*}
where we call $\mathbf{y}_i \in \mathbb{R}^d$ the embedding of the $i$th data point $\mathbf{x}_i$, with $i \in \{1, ..., m\}$. With both layers, the network computes
\begin{align}\label{eq:nnsvd}
f(\mathbf{x}_i) &= f'(\mathbf{x}_i) W_2 = (\mathbf{x}_i W_1) W_2 = \mathbf{y}_i W_2 = \mathbf{\hat r}_i,
\end{align}
the $n$-dimensional vector $\mathbf{\hat r}_i$. Expressed in matrix notation, given an input matrix $X \in \mathbb{R}^{m\times m}$ the network first computes an embedding matrix $Y \in \mathbb{R}^{m \times d}$ and from it the output ${\hat R} \in \mathbb{R}^{m \times n}$:
\begin{align*}
f(X) &= f'(X)W_2 = (XW_1)W_2 = YW_2 = {\hat R}.
\end{align*}
If the network is now trained (with backpropagation) to minimize the mean squared error of its output to a target matrix $R$, i.e.\vspace{-10pt}
\begin{align*}
\min\; \left\|R - f(X)\right\|_F^2,
\end{align*}
while we use as input to the network the identity matrix, i.e.,\ $X = I_m$, then, once the weights of the network have converged to a local optimum, $W_1W_2$ is a low rank approximation of the matrix $R \in \mathbb{R}^{m\times n}$ \cite{cichocki1992neuralsvd}.

When computing an SVD of a matrix, the eigenvectors stored in the matrices $U$ and $V$ are orthogonal (i.e.~$V^\top V = I_n$), which can be added as a further constraint to the cost function:
\begin{align*}
\min\; \left\|R - I_mW_1W_2\right\|_F^2 + \lambda \left\|I_dW_2W_2^\top - W_2W_2^\top\right\|_F^2,
\end{align*}
where $\lambda$ is a hyperparameter to control the strength of this regularization.\footnote{While this will encourage orthogonal rows in $W_2$, since the rows do not need to have unit length, the values on the diagonal of $W_2W_2^\top$ should not be penalized. This kind of regularization is usually only necessary if $d$ is chosen to be greater than the number of significant eigenvalues. Please note that the rows of $W_2$ are not necessarily ordered by the magnitude of the corresponding eigenvalues.}

Should $R$ contain missing values, then the error used in the backpropagation procedure to tune the network's parameters is only computed considering the available entries of the matrix. In this case especially it is advisable to additionally use other regularization techniques such as adding $\ell_2$ regularization terms to the cost function.

As the decomposition of a square symmetric matrix $S \in \mathbb{R}^{m\times m}$ into its eigenvalues and -vectors is a special case of an SVD (where $V = U$), the same NN can be used, only with an additional constraint to learn a symmetric factorization, i.e., to encourage $I_mW_1 = Y = W_2^\top$. This can be achieved with the cost function\vspace{-5pt}
\begin{align*}
\min\; \left\|S - I_mW_1W_2\right\|_F^2 + \lambda \left\|S - W_2^\top W_2\right\|_F^2,
\end{align*}
where, after convergence, $YW_2 \approx W_2^\top W_2 \approx YY^\top \approx S$. This also results in the same eigenvector based embedding $Y \in \mathbb{R}^{m \times d}$ as found by kernel PCA.

\subsection{Similarity Encoders}\label{sec:simec}
\begin{figure*}[!ht]
  \centering
      \includegraphics[width=0.68\linewidth]{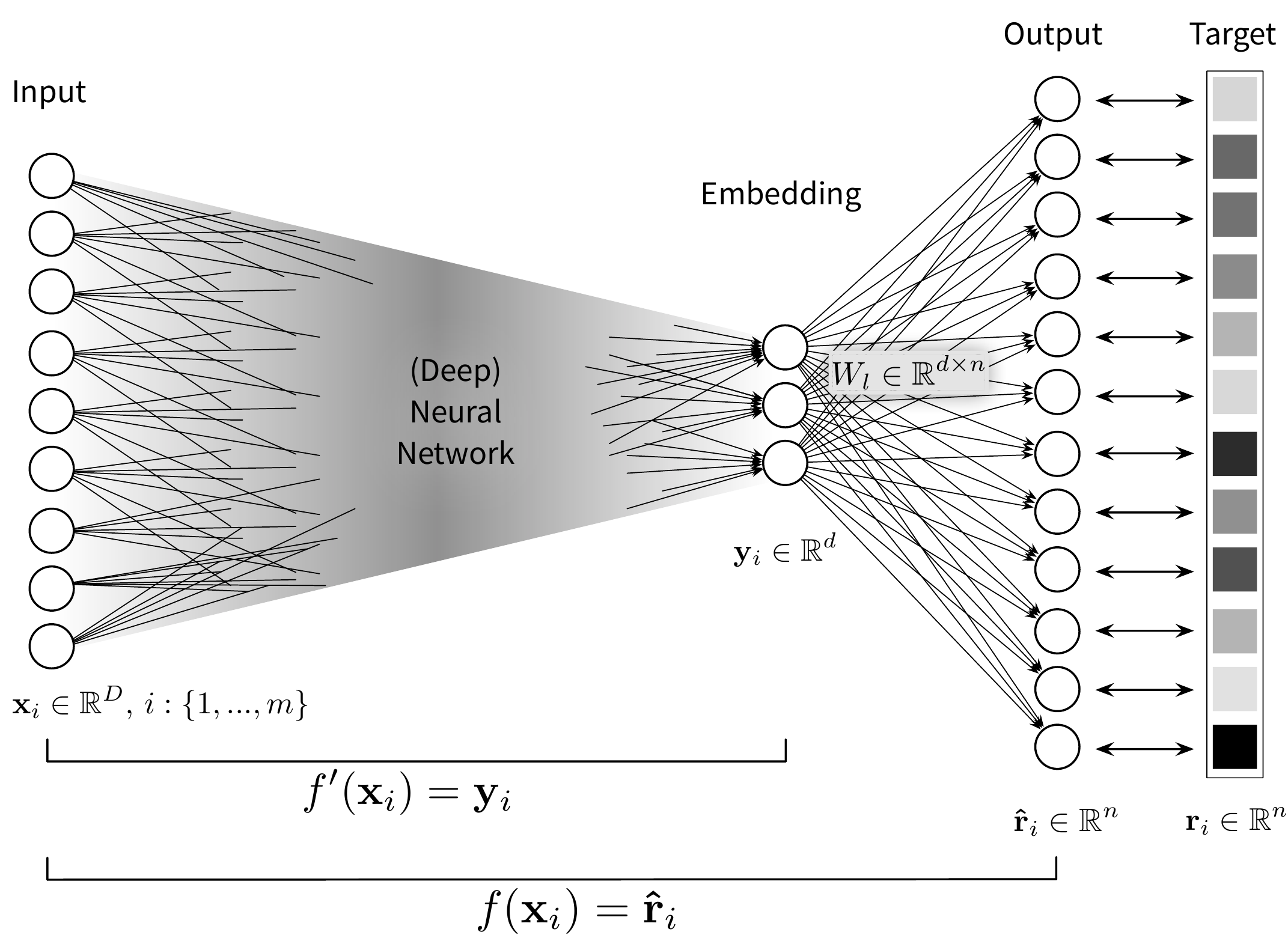}
  \caption{Similarity encoder (SimEc) architecture. A (deep) neural network, $f'(\mathbf{x}_i)$, is used to map the original feature vector $\mathbf{x}_i \in \mathbb{R}^{D}$ to an embedding $\mathbf{y}_i \in \mathbb{R}^{d}$. This embedding is then multiplied by another weight matrix $W_{l} \in \mathbb{R}^{d \times n}$, which corresponds to the last layer of the full SimEc network $f(\mathbf{x}_i)$, to compute $\mathbf{\hat r}_i \in \mathbb{R}^{n}$, i.e., the approximation of one row of the target matrix $R \in \mathbb{R}^{m \times n}$. After the SimEc is trained to minimize $\left\|R - f(X)\right\|_F^2$ given the full feature matrix $X \in \mathbb{R}^{m\times D}$, it then computes a rank $d$ approximation of $R$ as $f(X) = f'(X)W_l = YW_l$.}
  \label{fig:simec_arch}
\end{figure*}

Now that the factorization of a matrix $R$ or $S$ is expressed in terms of optimizing a neural network, this setup can be further extended to yield our SimEc architecture.
In particular, the first linear layer of the neural network, $f'(\mathbf{x}_i) = \mathbf{x}_i W_1 = \mathbf{y}_i$, can be replaced by any kind of (deep) neural network to map arbitrary feature vectors $\mathbf{x}_i \in \mathbb{R}^D \; \forall i \in \{1,...,m\}$ into the low dimensional embedding space (Fig.~\ref{fig:simec_arch}).
Equation~\eqref{eq:nnsvd} then becomes\vspace{-5pt}
\begin{align*}
f(\mathbf{x}_i) &= f'(\mathbf{x}_i)W_{l} = \mathbf{y}_iW_{l} = \mathbf{\hat r}_i,
\end{align*}
where again $\mathbf{y}_i \in \mathbb{R}^d$ is the embedding of the $i$th data point $\mathbf{x}_i$ and $W_l \in \mathbb{R}^{d \times n}$ is the weight matrix of the last (linear) layer of the full network $f(\mathbf{x}_i)$, while $f'(\mathbf{x}_i)$ could, for example, be a convolutional neural network (CNN) mapping images into the embedding space. This similarity encoder network is again trained to minimize $\left\|R - f(X)\right\|_F^2$, thereby learning the factorization $R \approx f'(X)W_l = YW_l$.

If the output of the SimEc should always be in a specific range, e.g., if the target matrix $R$ contains star ratings from 1 to 5, it may be beneficial to add an additional non-linearity after computing $f'(\mathbf{x}_i)W_{l}$ to ensure the predicted values are within this range. However, there should not be any non-linearity at the last layer of $f'(\mathbf{x}_i)$ as the embedding values $\mathbf{y}_i$ should be able to assume unconstrained values.

Regularization terms can again be added to the cost function as discussed before. However, it should be noted that the constraint to encourage a symmetric factorization of a similarity matrix $S$, i.e., the regularization term $\left\|S - W_l^\top W_l\right\|_F^2$, can significantly increase the computational complexity of the optimization procedure, as computing $W_l^\top W_l$ scales with $m^2$. However, in practice it is often enough to only train with a subsample of $S$ using $n \ll m$ targets, i.e., optimizing
\begin{align*}
\min\; \left\|S_{[:,:n]} - f'(X)W_l\right\|_F^2 + \lambda \left\|S_{[:n,:n]} - W_l^\top W_l\right\|_F^2,
\end{align*}
with $W_l \in \mathbb{R}^{d\times n}$ and $f(X) = {\hat S} \in \mathbb{R}^{m \times n}$, which greatly reduces the overall complexity and memory requirements of the training procedure. Even though the number of targets in the output is reduced, all $m$ training examples can still be used as input during training.

Instead of limiting the number of targets, it might also be worth considering whether it is necessary to enforce a symmetric factorization of $S$ (as $YW_l \approx W_l^\top W_l \approx YY^\top$) at all. If the SimEc only needs to predict the similarities between a new sample and the existing samples or even just between the existing samples themselves, e.g., to fill missing values in $S$, then the regularization term can in practice be ignored. The similarities between a new sample $\mathbf{x}_j$ and the training samples can then be computed as
\begin{align*}
f(\mathbf{x}_j) &= f'(\mathbf{x}_j)W_l = \mathbf{y}_jW_l = \mathbf{\hat s}_j \in \mathbb{R}^{m},
\end{align*}
instead of $f'(\mathbf{x}_j)f'(X)^\top = \mathbf{y}_jY^\top$.

A similar choice should be made when factorizing a rectangular matrix $R$. By default a SimEc only learns the mapping from one input feature space to the embedding space and then predicts the values of $R$ by multiplying this embedding with $W_l$. This is sufficient in many cases. For example, for an established social network site, thousands of pieces of new content are uploaded every second and at the same time older content becomes irrelevant, while the user base remains fairly constant. In such a scenario it might be sufficient to simply predict which users might be interested in a new piece of content, which can be done by using the full SimEc network to predict $f(\mathbf{x}_i) = \mathbf{\hat r}_i$ for some content feature vector $\mathbf{x}_i \in \mathbb{R}^D$. Nevertheless, it is also possible to train a second SimEc network to additionally project the set of $n$ users into the same embedding space as some $m$ items, thereby making it possible to predict ratings for both new items \emph{and} new users as the scalar product of their embedding vectors. For this, a SimEc network $f_1$ is first trained on one set of feature vectors $X_1 \in \mathbb{R}^{m \times D}$ to approximate $R$ (or a subset of it). After the training is complete, these feature vectors are projected into the embedding space to yield $f_1'(X_1) = Y_1 \in \mathbb{R}^{m \times d}$. Then, a second SimEc $f_2$ can be trained using the second set of feature vectors  $X_2 \in \mathbb{R}^{n \times P}$ to approximate $R^\top$ (or again a subset of it), only that in this case the weights of the last layer are kept fixed as $W_l = Y_1^\top$. Both SimEcs together then provide mapping functions for two different kinds of input feature vectors into the same embedding space such that $f_1'(X_1)f_2'(X_2)^\top = Y_1Y_2^\top \approx R$.

\subsubsection{Preserving non-metric similarities and predicting multiple pairwise relations at once}
Non-metric similarities are characterized by an eigenvalue spectrum with significant negative eigenvalues. Spectral embedding methods such as kPCA require positive semi-definite similarity matrices to compute the low dimensional embedding of the data and would in this case discard the information associated with the negative eigenvalues. However, Laub et al.~\cite{laub2004feature} have shown that this negative part of the eigenvalue spectrum can reveal interesting features in the data and therefore should not be ignored.

A non-metric similarity matrix $S$ is equal to the difference between two similarity matrices $S_1$ and $S_2$, where $S_1$ has the same $p$ positive eigenvalues as $S$, while the non-zero eigenvalues of $S_2$ correspond to the $q$ negative eigenvalues of $S$. Correspondingly, a factorization of $S$ into $YY^\top$ would need to capture the relation between $S_1$ and $S_2$, i.e.,
\begin{align*}
S &= S_1 - S_2 \approx YY^\top = Y_pY_p^\top - Y_qY_q^\top.
\end{align*}
However, the only way to get this negative part of the product $YY^\top$ would be for the values of $Y_q$ to be imaginary, which is generally not desirable for such embeddings.

With SimEcs it is nevertheless possible to approximate a non-metric similarity matrix $S$. Since during training $S$ is approximated as $f'(X)W_l = YW_l$ and not $YY^\top$, some parts of $Y$ and $W_l$ can have opposite signs, which makes it possible to not only approximate $S_1$ but also $(-S_2)$. In this case the regularization term $\left\|S - W_2^\top W_2\right\|_F^2$ would be counterproductive.\footnote{It should be noted that a $d$-dimensional SimEc embedding generally captures the information associated with the $d$ eigenvalues with the largest absolute values; should the magnitude of the largest negative eigenvalue be smaller than the first $d$ positive values, then this information will still be ignored.}

SimEc can also be trained explicitly to preserve the information provided by multiple similarity matrices $S_1, ..., S_k$. The easiest way to do this is to simply compute the average of these similarity matrices and then train a SimEc as before on this averaged $S$. However, because SimEcs preserve the information associated with the $d$ largest eigenvalues, the embedding only captures all $k$ similarities if the largest eigenvalues of the $k$ similarity matrices are equal. Therefore, before computing their average, the similarity matrices should first be normalized by dividing them each by their respective largest eigenvalue.

If the focus is not on the similarity preserving embedding itself, but rather it is important to accurately predict multiple similarities or other pairwise relations at the same time, then the SimEc network can be extended to have multiple last layers, i.e., by choosing $W_l \in \mathbb{R}^{d \times n \times k}$ a SimEc can compute
\begin{align*}
f(X) &= f'(X)W_l = YW_l = {\hat R} \in \mathbb{R}^{m \times n \times k}.
\end{align*}
Similarly, in addition to a last layer $W_l$, the SimEc network can also be extended by a mirrored version of $f'(\mathbf{x}_i)$, thereby adding a decoder part to the network, which can be used to compute the pre-image of an embedding like in the deep kernelized autoencoder networks \cite{kampffmeyer2017deep}.

\section{Experiments \& Results}\label{sec:exp}
In the following, we demonstrate that SimEc can learn a mapping from an original input feature space into a similarity preserving embedding space, even if the target similarities were not computed from the original feature vectors. Furthermore, we discuss the influence of regularization and the number of targets on the embedding quality, as well as show that SimEc can create a faithful embedding even if the target similarity matrix contains over $90\%$ missing values. Finally, we demonstrate that SimEc can predict non-metric similarities and multiple similarities at once.

As SimEcs simultaneously factorize a similarity matrix and learn a mapping into the similarity preserving embedding space, the most appropriate method to compare a SimEc's performance with is the combination of the eigendecomposition of $S$, to get optimal similarity preserving embeddings, and an additional regression model, trained to learn the mapping from the original feature space to the embedding space. As the embeddings produced by the regression model will at most be as good as the original embeddings created by decomposing $S$ \cite{carreira2015fast}, in most experiments we only report the optimal performance achieved by the eigendecomposition as a reference.

Further details as well as the code to replicate these experiments and more is available online \cite{simec_github}.

\subsection{Dataset}
All experiments are performed on subsets of the MNIST dataset, which contains $28\times 28$ pixel images depicting handwritten digits.
For the first set of experiments, we randomly subsampled 10k images from all classes, of which 80\% are assigned to the training set and the remaining 20\% to the test set. For the second set of experiments, we randomly subsampled 5k images depicting zeros and sevens and we refer to this as the ``MNIST 0/7'' dataset.

As input feature vectors we use the 784 pixel values of each image, which we normalize by their maximum value and center to have zero mean. The respective target similarity matrices were also centered (as it is being done for kPCA as well \cite{muller2001introduction}) and, if necessary, normalized to be in the range $[-1, 1]$.

\setcounter{figure}{3}
\begin{figure*}[!ht]
  \centering
      \includegraphics[height=3.93cm]{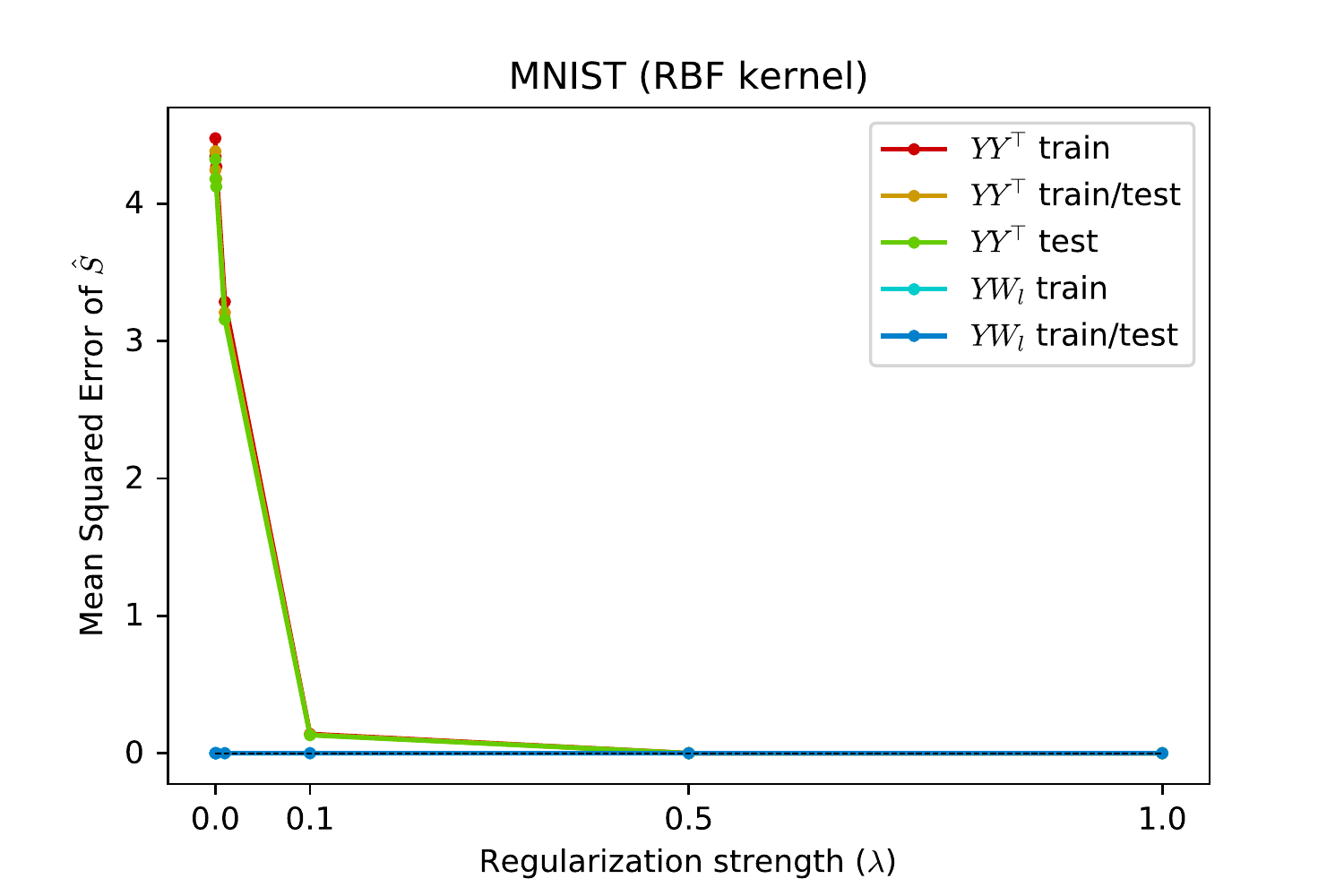}
      \includegraphics[height=3.93cm]{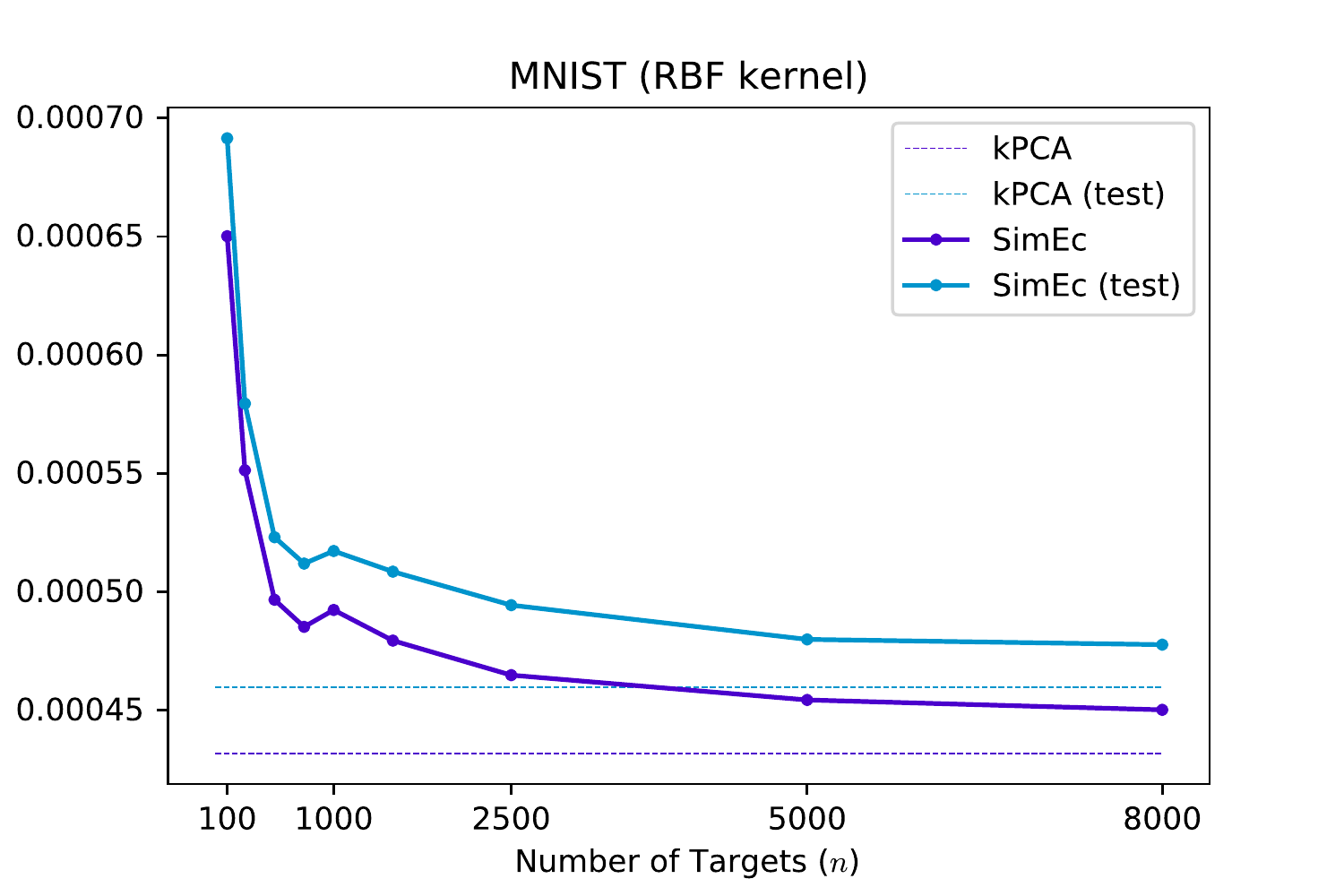}
      \includegraphics[height=3.93cm]{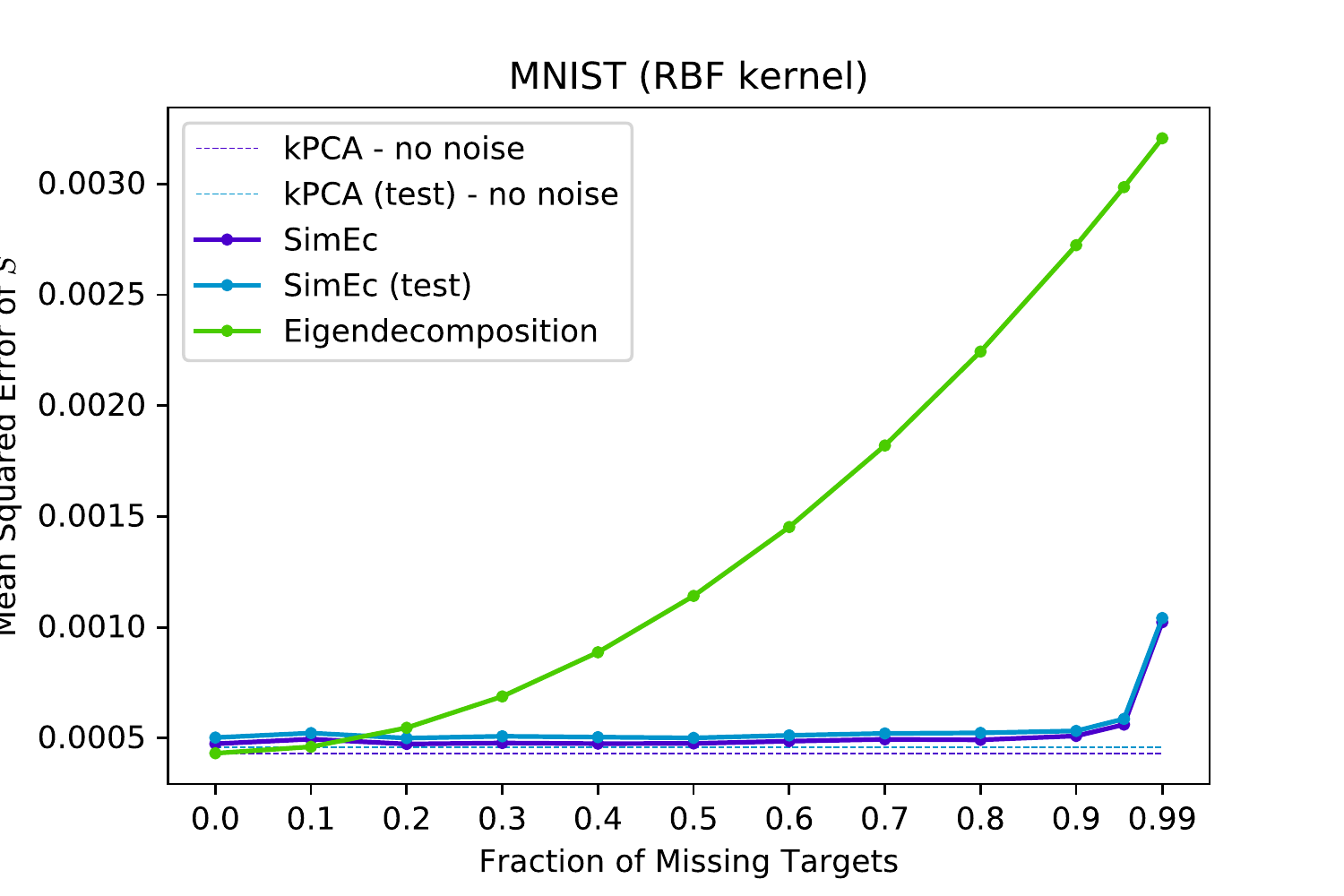}
  \caption{\emph{Left:} Importance of the regularization term $\lambda \left\|S - W_l^\top W_l\right\|_F^2$ to ensure not only the output of the SimEc network, $YW_l$, approximates the target similarity matrix, but also the dot product of the embedding vectors, $YY^\top$. With $YW_l$ it is only possible to predict the similarities between new samples and those used for training the network, while with $YY^\top$ the similarities between new test samples can be computed as well. \emph{Middle:} Even if only a fraction of targets is used for training, the mean squared error between $YY^\top$ and $S$ is close to the optimal error achieved by kernel PCA. \emph{Right:} Influence of missing values in the target similarity matrix. Kernel PCA computed on the full matrix again serves as the optimal reference error, while the green curve depicts the error achieved by computing the eigendecomposition of the matrix where the missing values were filled with the mean of the matrix.}
  \label{fig:targets}
\end{figure*}
\subsection{Mapping into a similarity preserving embedding space}
To demonstrate that SimEc can learn the connection between data points' feature vectors and an unrelated target similarity matrix $S$, we compute pairwise similarities between the MNIST images based on their class labels. This similarity matrix is 1 for a pair of images depicting the same digit and 0 elsewhere. With increasing embedding dimensionality $d$, the mean squared error between the target similarity matrix $S$ and its approximation ${\hat S}$, computed as the dot product of the embedding vectors, $YY^\top$, should decrease. The eigendecomposition of $S$ provides the optimal similarity preserving embeddings. However, this does not provide a mapping from the original input feature space to the embedding space to compute OOS solutions, as for new test samples the class based similarities are not available.
\setcounter{figure}{2}
\begin{figure}[!ht]
  \centering
      \includegraphics[width=0.9\linewidth]{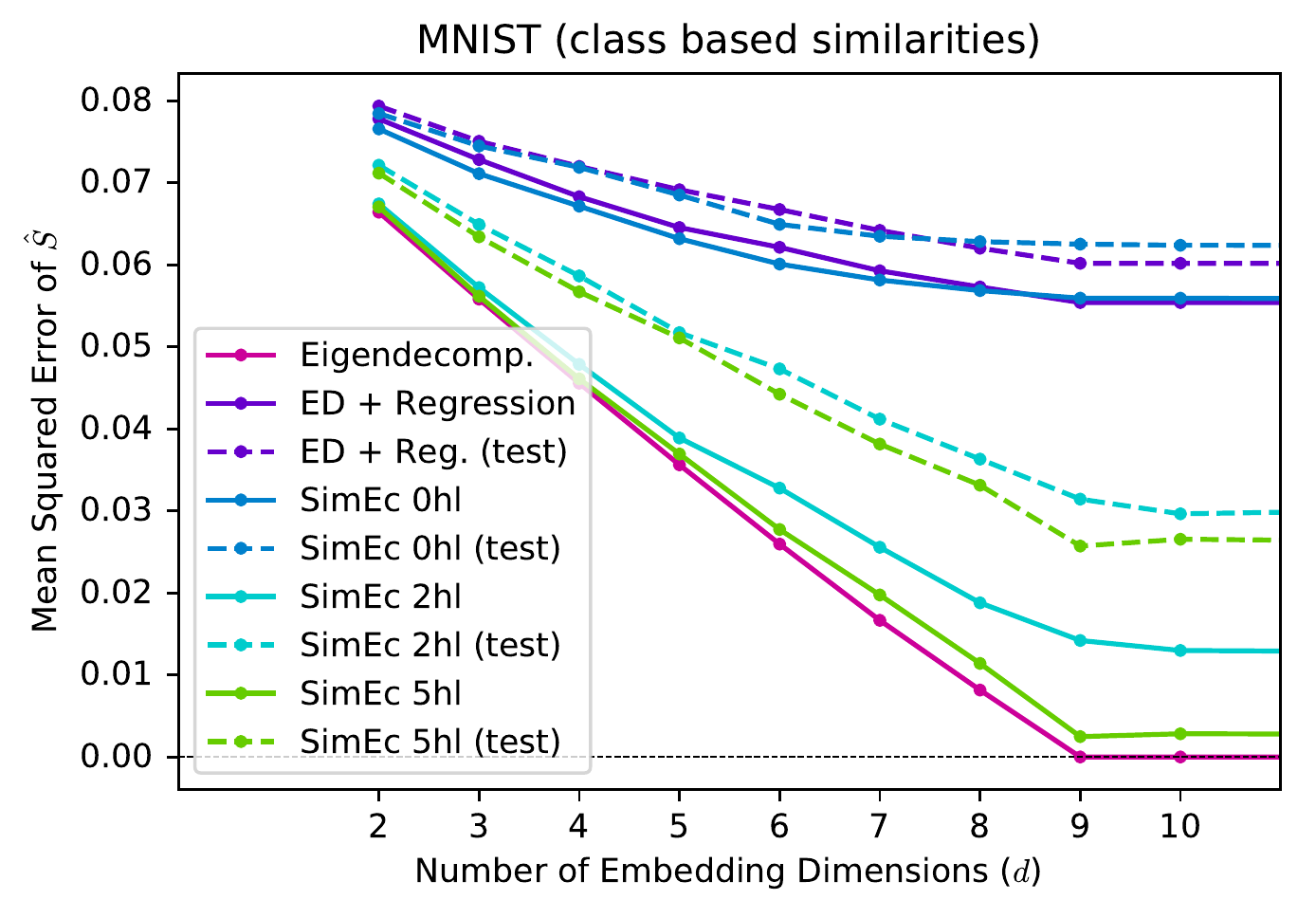}
  \caption{Mean squared errors between the target similarity matrix $S$ and its approximation ${\hat S}$, computed as the dot product of the embedding vectors, $YY^\top$, with increasing embedding dimensionality $d$.}
  \label{fig:classmse}
\end{figure}
\setcounter{figure}{4}
As shown in Fig.~\ref{fig:classmse}, the embeddings produced by a linear SimEc, where $f'(\mathbf{x}_i)$ consists of only a single linear layer mapping the input vectors into the embedding space, are comparable to those of a linear ridge regression model that learned the connection between the feature vectors and the embeddings produced by the eigendecomposition of $S$. By using a SimEc with a deeper NN $f'(\mathbf{x}_i)$ with several non-linear hidden layers to map the feature vectors into the embedding space, the error of the approximation gets very close to that of the eigendecomposition.

\subsection{Of hyperparameters and missing values}
Next, we investigate the influence of hyperparameter choices and missing values in the target similarity matrix. For this, a SimEc with one additional hidden layer is trained to create ten dimensional embeddings to approximate an RBF kernel matrix. Corresponding embeddings created with kernel PCA serve as a reference.

First, we analyze the influence of the regularization term $\lambda \left\|S - W_l^\top W_l\right\|_F^2$ (Fig.~\ref{fig:targets} left panel). While the output of the SimEc network, $YW_l$, always faithfully approximates the target similarities, the dot product of the embedding vectors, $YY^\top$, only achieves similar accuracies when a symmetric factorization of $S$ is enforced.

As we discussed before, this regularization dramatically increases the computational complexity and memory requirements of the training procedure, as it scales quadratically with the output dimensionality. However, often only a fraction of the targets is required for $YY^\top$ to approximate $S$ reasonably well (Fig.~\ref{fig:targets} middle).

As pairwise data can be expensive to collect or be systematically unavailable (e.g.~in movie ratings), target matrices will often contain many missing values. An exact eigendecomposition of a matrix with missing values can not be computed, and instead these entries in the matrix need to be filled, e.g., by the mean of the given targets. However, this results in an almost linear increase in the mean squared error between the full target matrix and the approximation computed as $YY^\top$ (Fig.~\ref{fig:targets} right panel). With the embeddings created with SimEc, on the other hand, the target similarities can be faithfully approximated even if the target matrix contains over $90\%$ missing values.

\subsection{Predicting non-metric similarities and more}
\begin{figure}[!ht]
  \centering
      \includegraphics[width=\linewidth]{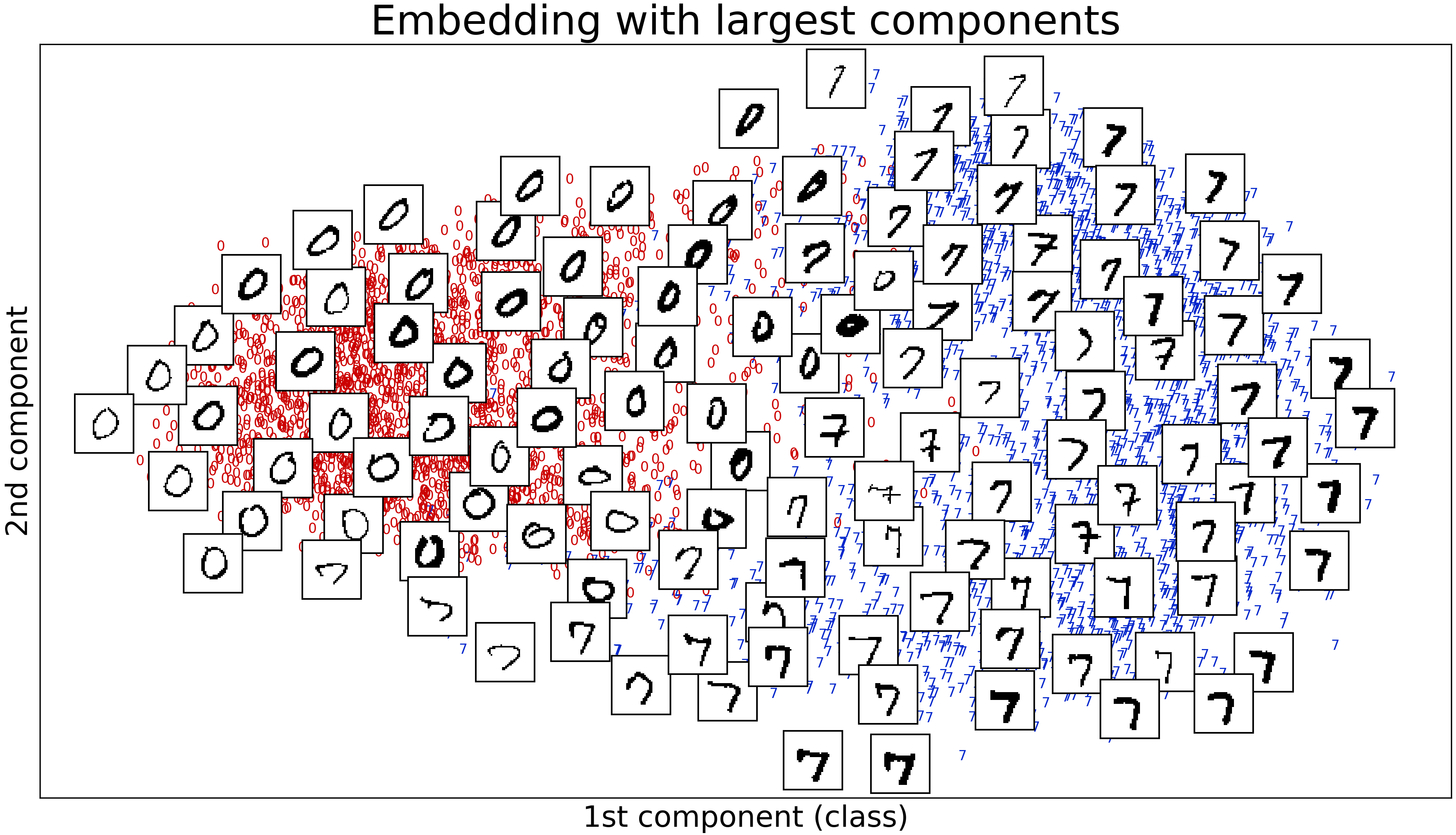}\\
      \includegraphics[width=\linewidth]{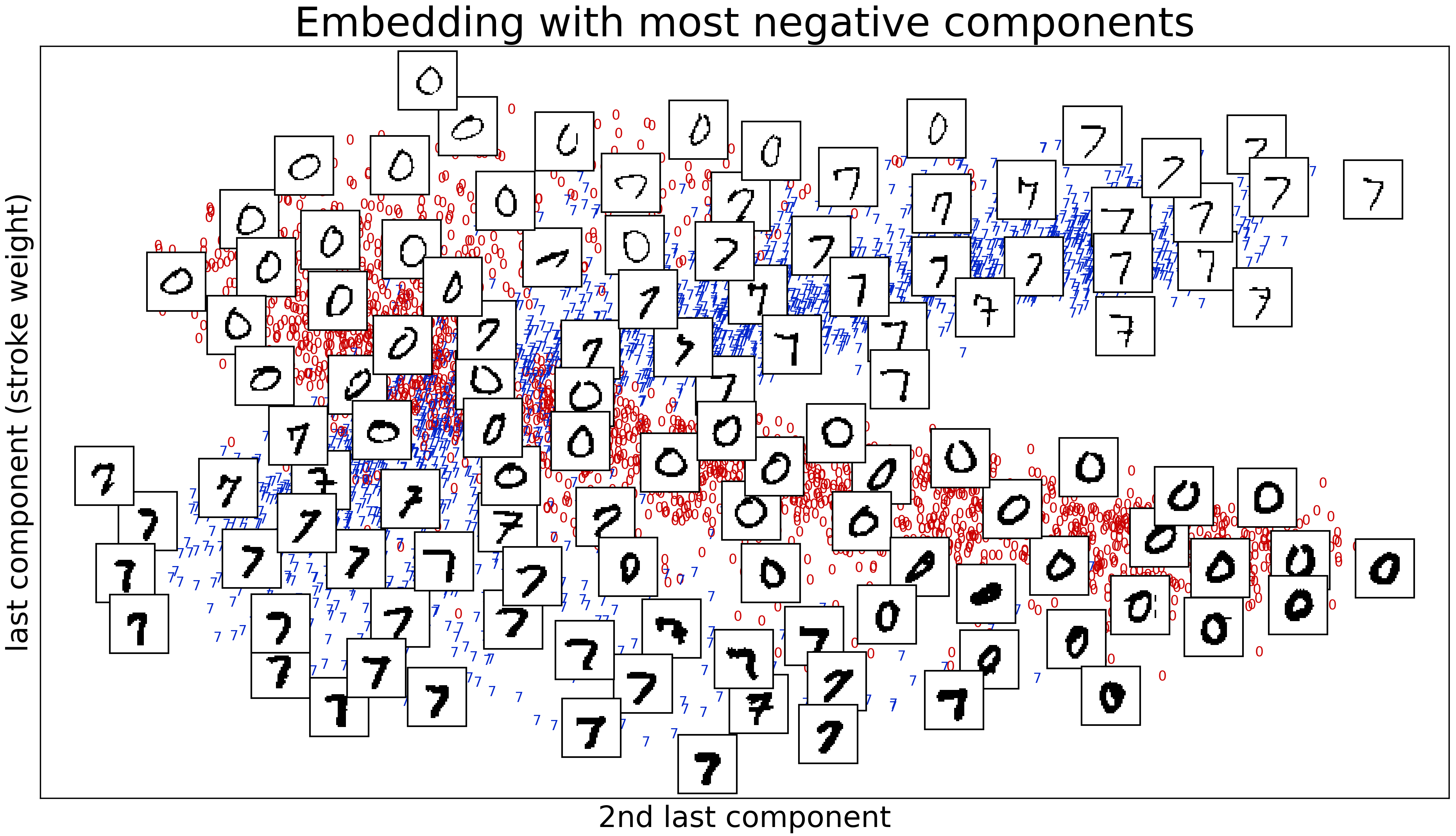}\\
      \includegraphics[width=\linewidth]{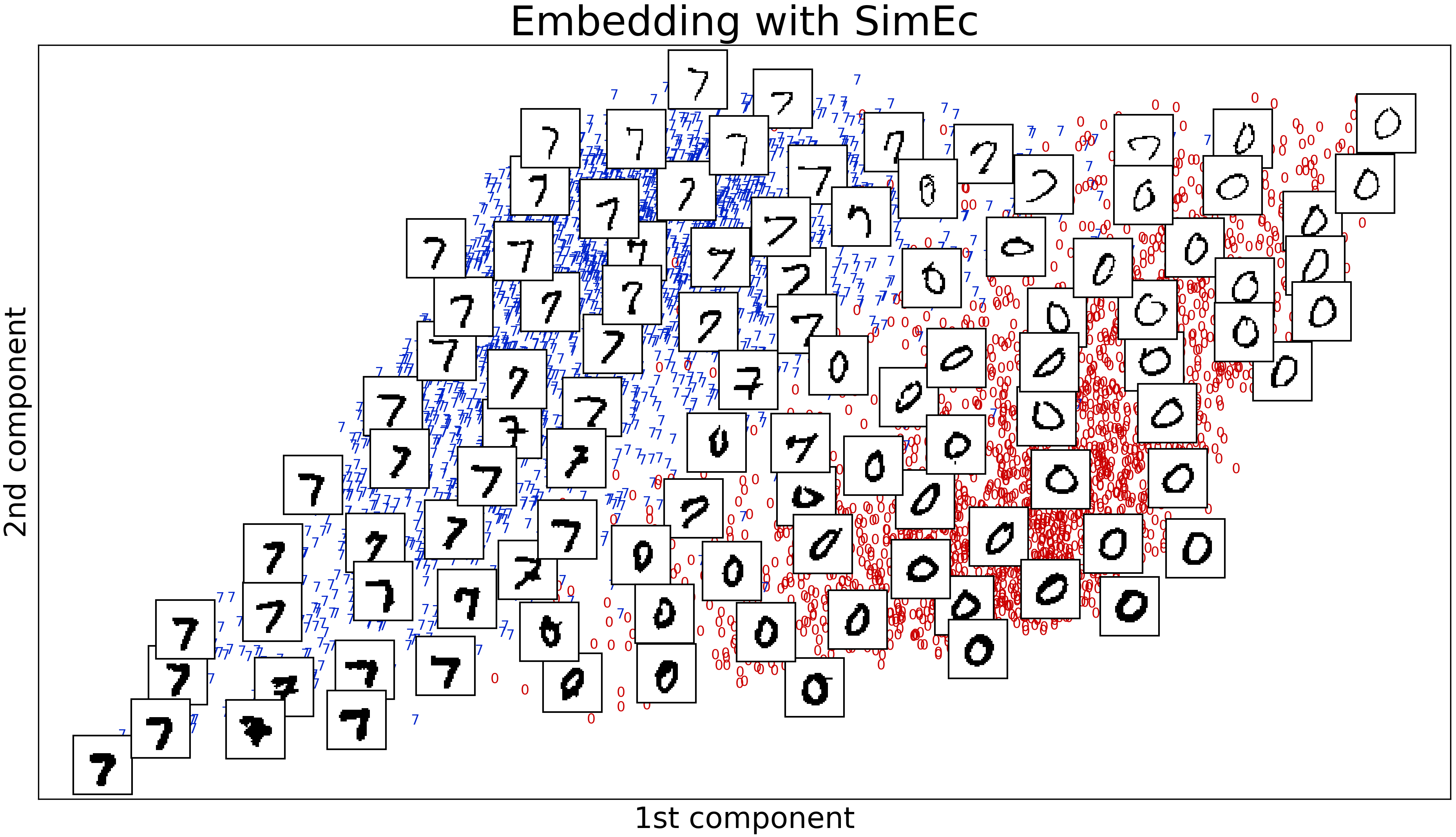}
  \caption{Embedding of the MNIST 0/7 dataset based on the largest \emph{(top)} and most negative \emph{(middle)} eigenvalues of the Simpson similarity matrix, as well as a SimEc embedding of dimensionality $d=2$ based on the sum of the similarity matrices associated with the largest and most negative eigenvalues \emph{(bottom)}.}
  \label{fig:nonmetric}
\end{figure}

\begin{figure}[!ht]
  \centering
      \includegraphics[width=0.9\linewidth]{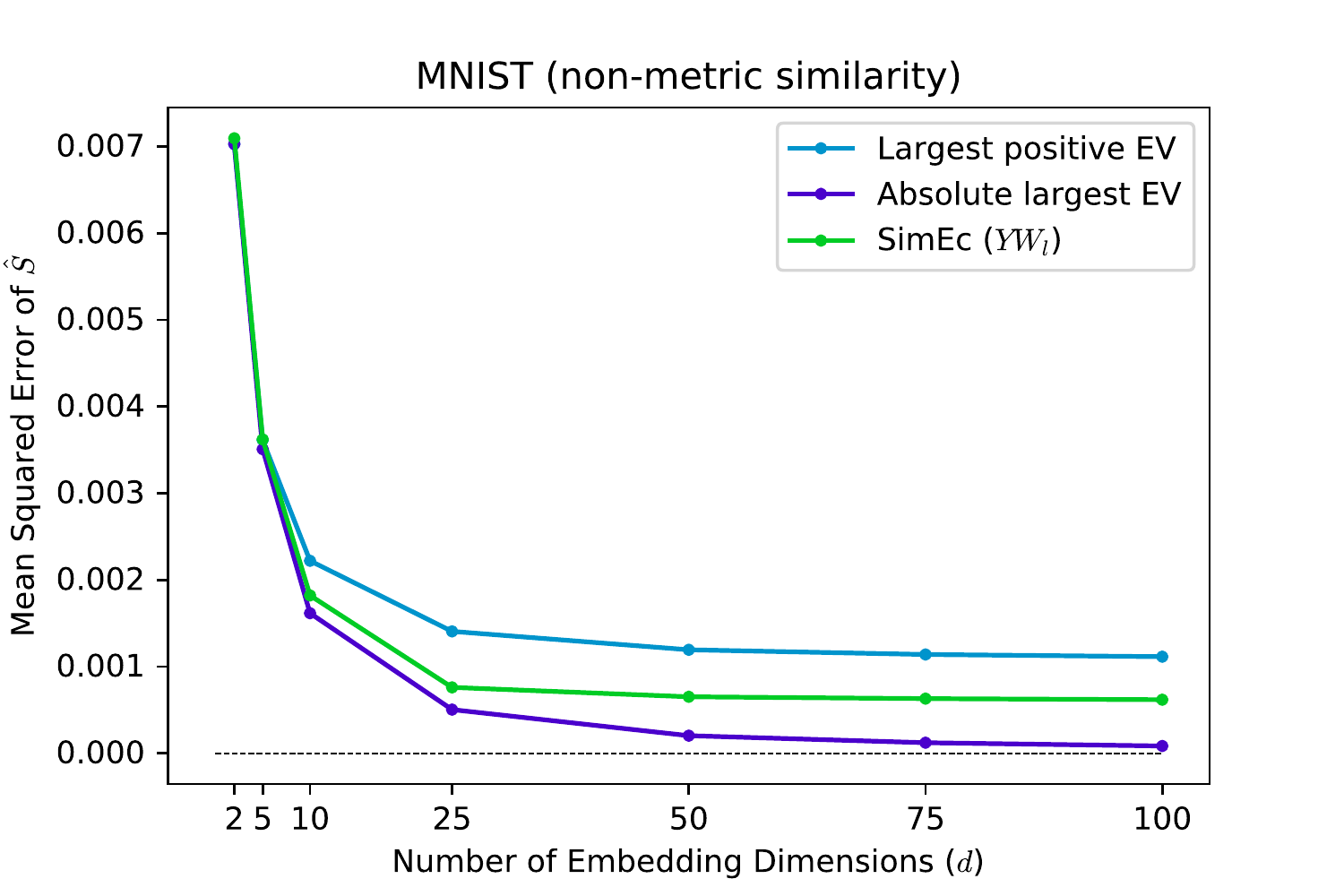}
  \caption{Mean squared errors of the non-metric similarity matrix $S$ and the dot product of the embeddings based on the largest positive eigenvalues, the embeddings based on the largest absolute eigenvalues (where dimensions associated with negative eigenvalues were cast as imaginary numbers), and the prediction of $S$ with a SimEc as $YW_l$.}
  \label{fig:nonmetric_mse}
\end{figure}

In the following experiments we demonstrate that SimEc can predict non-metric similarities and multiple similarities at once. For this we use the MNIST 0/7 dataset and compute the target similarity matrix $S$ using the Simpson similarity score on binarized feature vectors:
\begin{align*}
S_{ij} = {\# \{\text{pixels that are black in both $i$ and $j$}\} \over \min\{\# \{\text{black pixels in $i$}\}, \# \{\text{black pixels in $j$}\}\}} .
\end{align*}
As previously shown by Laub et al.~\cite{laub2004feature}, the eigenvalue spectrum of this matrix contains significant negative eigenvalues and embeddings based on the corresponding eigenvectors reveal interesting features. While the embedding based on the largest eigenvalues separates the data points by class (Fig.~\ref{fig:nonmetric} top), an embedding based on the most negative eigenvalues sorts the images by stroke weight (Fig.~\ref{fig:nonmetric} middle).
SimEc are able to create embeddings based on non-metric similarities as well. While the embedding learned by a SimEc (with one hidden layer) captures the features associated with the negative eigenvalues, their dot product would not optimally approximate $S$, as for this the dimensions associated with the negative eigenvalues would have to be imaginary. However, by computing ${\hat S} = YW_l$ the non-metric similarities can be predicted quite well (Fig.~\ref{fig:nonmetric_mse}), with errors closer to those of the embeddings based on both positive and negative eigenvalues instead of those of the embeddings based only on the largest positive eigenvalues (i.e.~a regular kPCA embedding).
\begin{figure*}[!ht]
  \centering
      \includegraphics[height=4.6cm]{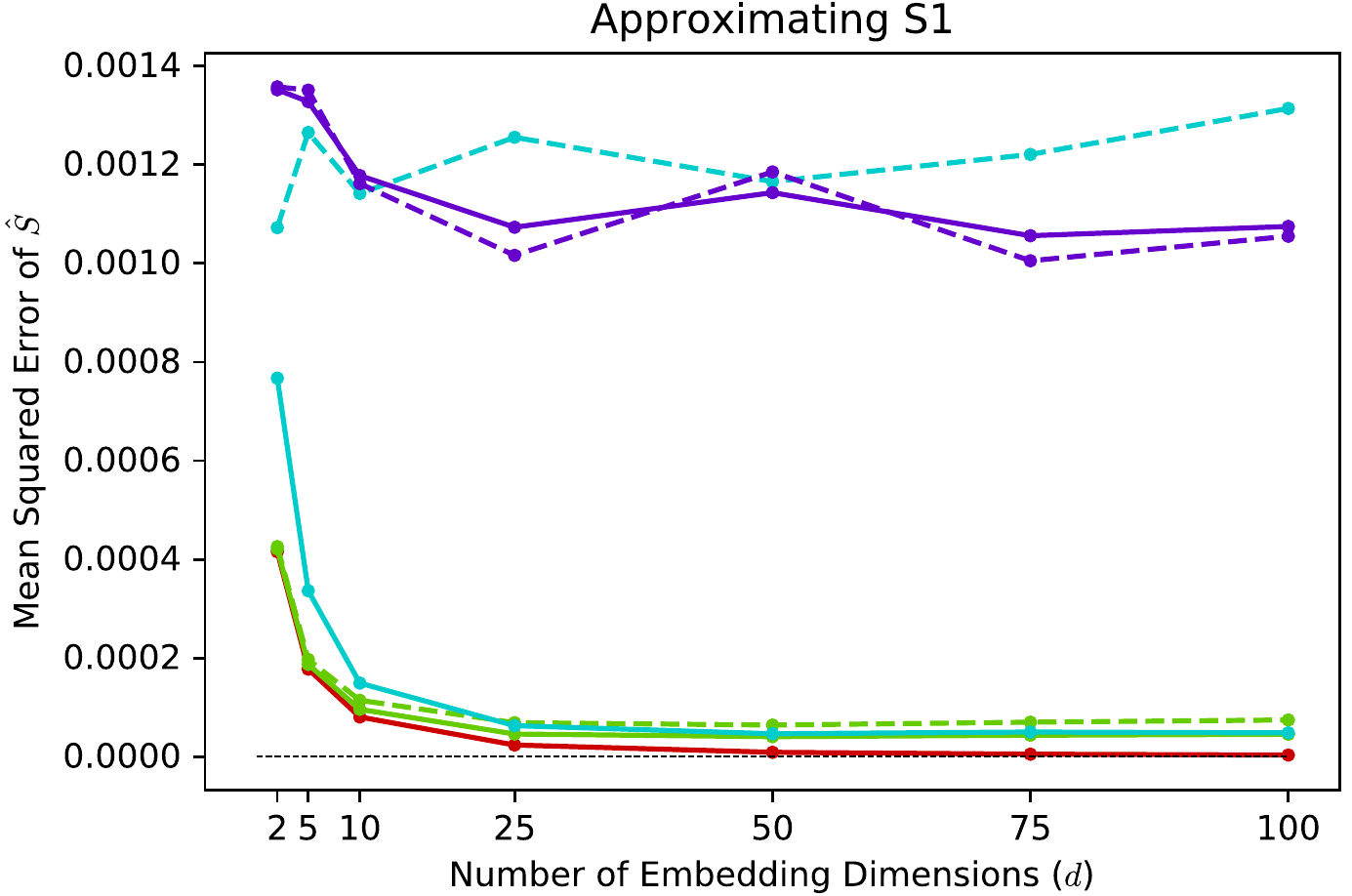}
      \includegraphics[height=4.6cm]{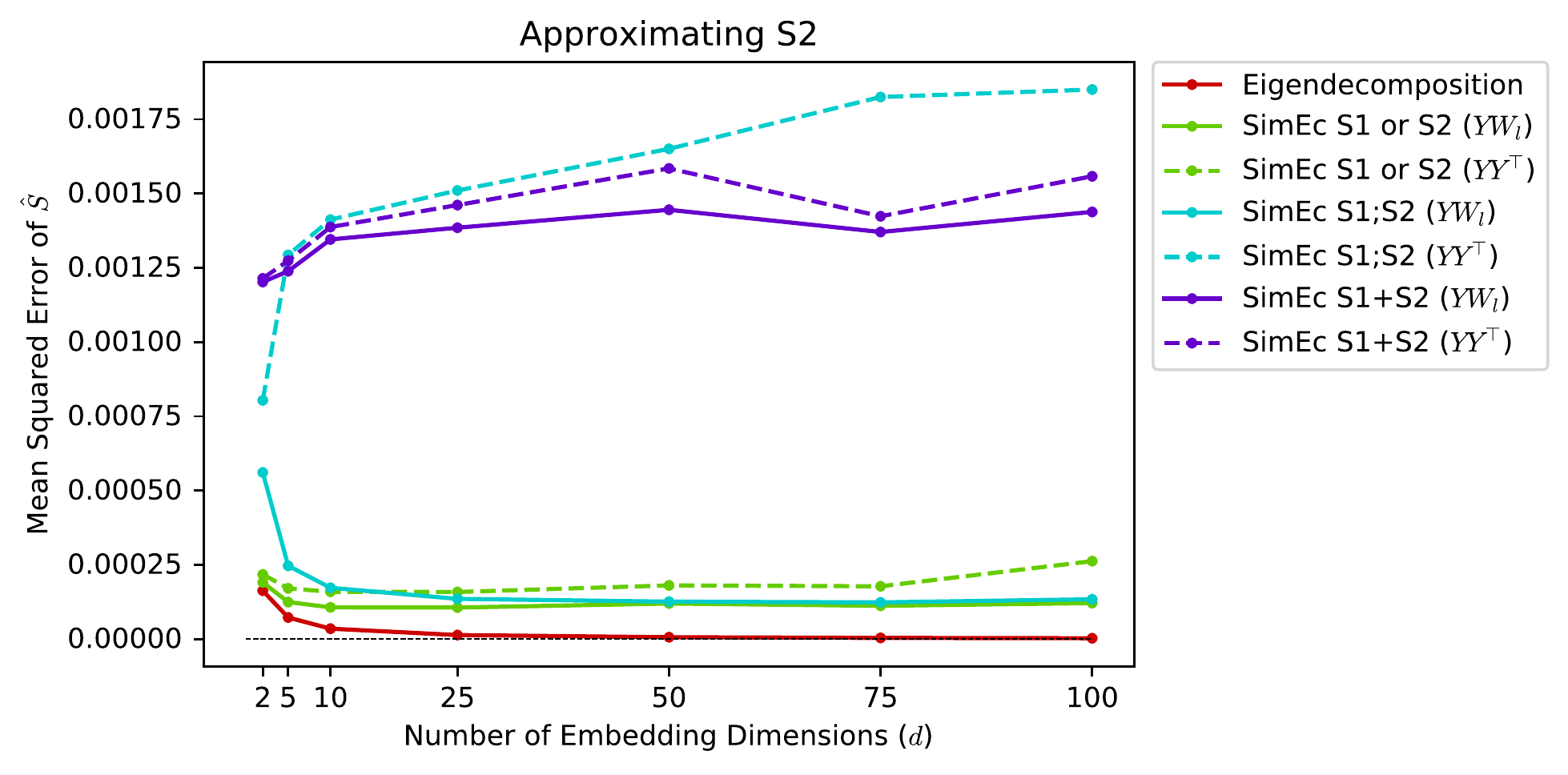}
  \caption{Mean squared errors when approximating either $S_1$ \emph{(left)} or $S_2$ \emph{(right)}. The eigendecomposition of the respective matrix yields the optimal similarity preserving embedding. Depicted in green are the errors achieved with a SimEc trained to approximate either $S_1$ or $S_2$ alone; shown in cyan are the errors achieved with a SimEc trained to approximate the tensor containing the stacked matrices $S_1$ and $S_2$; while the purple curves show the errors achieved with a SimEc trained to approximate the matrix $S_1+S_2$. Continuous lines depict the prediction of $S$ as $YW_l$, while dashed lines correspond to the approximation as $YY^\top$.}
  \label{fig:nonmetric_s1s2}
\end{figure*}

As discussed in the previous section, the non-metric similarity matrix can be decomposed as $S = S_1 - S_2$, where $S_1$ and $S_2$ can be computed as the dot product of the embeddings based on positive and negative eigenvalues respectively. Besides preserving features corresponding to both parts of the eigenvalue spectrum, SimEc can also be used to directly predict these two similarity matrices simultaneously. This can either be done by computing a new similarity matrix as $S_1+S_2$ (Fig.~\ref{fig:nonmetric} bottom), or by stacking the two matrices, thereby creating a tensor $\in \mathbb{R}^{m\times m\times 2}$. To preserve the information present in both similarity matrices to an equal extent, $S_1$ and $S_2$ first have to be normalized by their respective largest eigenvalue, as SimEc generally learn embeddings based on the overall largest eigenvalues.
Unsurprisingly, the mean squared error between either $S_1$ or $S_2$ and ${\hat S}$ computed with a SimEc trained to approximate $S_1+S_2$ is worse than that of a SimEc trained specifically to approximate either $S_1$ or $S_2$ alone (Fig.~\ref{fig:nonmetric_s1s2}). The dot product of the embedding vectors $YY^\top$ of a SimEc trained to approximate the tensor containing the stacked matrices $S_1$ and $S_2$ also results in an error comparable to that of the $S_1+S_2$ SimEc, because a single embedding contains the information about both similarity matrices here as well. However, the prediction of the individual similarity matrices in the tensor as $YW_l$ yields errors as low as the prediction of the SimEc trained to approximate only one of the matrices, because the last dimension of the tensor $W_l$ contains information specific to either one of the similarity matrices.

\section{Discussion}
Representing intrinsically complex structured data is an ubiquitous challenge in machine learning. While spectral methods such as kernel PCA provide optimal similarity preserving embeddings by computing the eigendecomposition of a similarity matrix, they are unable to produce OOS solutions for new test samples if their similarity to the original training examples can not be computed. Neural network based methods provide a mapping function from an original input feature space to the embedding space and can therefore also approximate the pairwise relations between new data points. However, existing methods were not designed to predict non-metric similarities or multiple pairwise relations simultaneously.

SimEc are a novel neural network architecture constructed for simultaneously learning a mapping from an original input feature space into a similarity preserving embedding space while factorizing a target matrix with pairwise relations. As we have demonstrated in multiple experiments, SimEc can provide OOS solutions even if the target similarities were obtained by an unknown process such as human ratings, they can efficiently handle missing values in the target matrix, and in addition they are able to predict non-metric similarities as well as multiple similarities at once.

While so far we mainly studied SimEcs based on fairly simple feed-forward neural networks, it appears promising to consider also deeper NN and more elaborate architectures, such as CNNs, for the initial mapping step to the embedding space. In this manner, hierarchical structures in complex data could be better reflected. Note furthermore that prior knowledge as well as more general error functions could be employed to tailor the embedding to the given targets.

In this paper we focused on using SimEc to predict pairwise similarities, but further application scenarios involving other pairwise relations between data points should be explored. For example, it has already been shown that a variant of SimEcs, called \emph{context encoders} (ConEc) \cite{horn2017conecRepL4NLP} learn meaningful word embeddings by extending the word2vec model \cite{mikolov2013efficient,mikolov2013distributed} for words with multiple meanings as well as to create out-of-vocabulary embeddings. The SimEc framework could also improve recommender systems or drug-protein interaction predictions and be interesting for usage in the sciences e.g., psychophysics \cite{laub2007inducing}, human quality judgment experiments \cite{bosse2018deep}, or materials discovery \cite{schutt2017quantum,schutt2018schnet}.

Furthermore, future work will aim to interpret the predictions made by SimEc using layer-wise relevance propagation \cite{arras2017relevant,bach2015pixel,kindermans2017patternnet,montavon2017explaining}.

\begin{acknowledgements}
We would like to thank Antje Relitz, Maximilian Alber, and Christoph Hartmann for their helpful comments on earlier versions of this manuscript.\\ Franziska Horn acknowledges funding from the Elsa-Neumann scholarship from the universities of Berlin. This work was supported by the Federal Ministry of Education and Research (BMBF) for the Berlin Big Data Center BBDC (No.\ 01IS14013A). Additional support was provided by the BK21 program funded by the Korean National Research Foundation Grant (No.\ 2012-005741) and the Institute for Information \& Communications Technology Promotion (IITP) grant funded by the Korea government (No.\ 2017-0-00451).
\end{acknowledgements}



\end{document}